\documentclass[journal]{IEEEtran}

\usepackage{cite}

\usepackage{amsmath}

\usepackage{array}

\usepackage{times}
\usepackage{epsfig}
\usepackage{graphicx}
\usepackage{amsmath}
\usepackage{amssymb}

\newcommand{\rb}{{\mathbf r}}
\newcommand{\Rd}{{\mathbb R}}

\usepackage{multirow, array}
\newcolumntype{P}[1]{>{\centering\arraybackslash}p{#1}}
\newcolumntype{M}[1]{>{\centering\arraybackslash}m{#1}}
\usepackage{caption}
\usepackage{algorithm}
\usepackage[noend]{algpseudocode}
\usepackage{xcolor}

\begin{document}
%
%
%
%

\title{{Mumford-–Shah Loss Functional for Image Segmentation with Deep Learning}}

\author{Boah~Kim
        and~Jong~Chul~Ye,~\IEEEmembership{Senior~Member,~IEEE}
\thanks{This work was supported by National Research Foundation of Korea under Grant NRF-2016R1A2B3008104.
This work is also supported by Industrial Strategic technology development program (10072064, Development of Novel Artificial Intelligence Technologies To Assist Imaging Diagnosis of Pulmonary, Hepatic, and Cardiac Disease and Their Integration into Commercial Clinical PACS Platforms) funded by the Ministry of Trade Industry and Energy (MI, Korea).}%
\thanks{B. Kim and J.C. Ye are with the Department of Bio and Brain Engineering, Korea Advanced Institute of Science and Technology (KAIST), Daejeon 34141, Republic of Korea (email: \{boahkim, jong.ye\}@kaist.ac.kr).}
}%
\maketitle

\begin{abstract}
Recent state-of-the-art image segmentation algorithms are mostly based on deep neural networks, thanks to their high performance and fast computation time. However, these methods are usually trained in a supervised manner, which requires large number of high quality ground-truth segmentation masks. On the other hand, classical image segmentation approaches such as level-set methods are formulated in a self-supervised manner by minimizing energy functions such as Mumford-Shah functional, so they are still useful to help generation of segmentation masks without labels. Unfortunately, these algorithms are usually computationally expensive and often have limitation in semantic segmentation. In this paper, we propose a novel loss function based on Mumford-Shah functional that can be used in deep-learning based image segmentation without or with small labeled data. This loss function is based on the observation that the softmax layer of deep neural networks has striking similarity to the characteristic function in the Mumford-Shah functional. We show that the new loss function enables semi-supervised and unsupervised segmentation. In addition, our loss function can be also used as a regularized function to enhance supervised semantic segmentation algorithms. Experimental results on multiple datasets demonstrate the effectiveness of the proposed method.
\end{abstract}


%
\IEEEpeerreviewmaketitle

\section{Introduction}

\IEEEPARstart{I}{mage} segmentation is to assign a label to every pixel in the image. Since image segmentation techniques {are required} for various applications such as object detection \cite{leibe2008robust} and medical image analysis \cite{lee2010segmentation, salih2005brain, cernazanu2013segmentation}, the segmentation study has been one of the core topics in computer vision. 

Recently, deep learning approaches for image segmentation have been developed and successfully demonstrated the state-of-the-art performance \cite{chen2018deeplab, kampffmeyer2016semantic}. A majority of the segmentation works are based on convolutional neural networks (CNN), which are usually trained in a supervised manner that requires large amount of data with high quality pixel-wise labels \cite{badrinarayanan2015segnet, noh2015learning, long2015fully, ronneberger2015u}. However, generating segmentation masks is time consuming and can be difficult in certain domains. To overcome this issue, weakly/semi-supervised image segmentation methods have been studied in these days. These methods are often under the assumption that training data have various supervisory annotations such as image-level labels \cite{hong2015decoupled, papandreou2015weakly, pinheiro2015weakly, qi2016augmented} and bounding-box labels \cite{dai2015boxsup,afshari2019weakly} or a small number of strongly annotated data.

On the other hand, classical variational image segmentation methods are typically implemented in an unsupervised manner, by minimizing a specific energy function {such as Mumford-Shah functional \cite{mumford1989optimal}} so that it can cluster image pixels into several classes \cite{mumford1989optimal,vese2002multiphase,kumarcolor, shi2000normalized, boykov2001interactive}. Since these methods produce pixel-wise prediction without ground-truth, they have extensively used for classical medical image segmentation \cite{mumford1989optimal,vese2002multiphase,kumarcolor, shi2000normalized, boykov2001interactive,liu2006automatic, duan2010coupled}. Unfortunately, these algorithms are usually computationally expensive and have limitation in semantic image segmentation, so recent segmentation studies have been mainly focused on convolutional neural network (CNN) approaches which are seemingly disconnected from classical approaches.

One of the most important contributions of this paper is, therefore, to show that the classical wisdom is indeed useful to improve the performance of the CNN-based image segmentation, specifically for the unfavorable situation where the training data is not sufficient. In particular, inspired by the {classical variational segmentation approaches}, here we propose a novel {loss function by employing classical {Mumford-Shah functional \cite{mumford1989optimal}}} that can be easily employed in CNN-based segmentation algorithms. Because the {Mumford-Shah functional} is based on the pixel similarities, the new loss can exploit the complementary information to the semantic information used in the existing CNN-based segmentation algorithms. Specifically, for a supervised learning setting, our loss function can be used as a novel regularized function to enhance the performance of the neural network by considering pixel similarities. Also, as the Mumford-Shah functional can be regarded as a self-supervised loss, we can easily obtain CNN-based semi-supervised or unsupervised learning without ground-truth labels or bounding box annotations. Furthermore, in the presence of intensity inhomogeneities, our Mumford-Shah functional for unsupervised segmentation can be easily modified to include bias field correction.

Recall that the minimization of Mumford-Shah functional has been the key topic in classical image segmentation so that various methods such as curve-evolution \cite{tsai2001curve,li2007implicit}, level set methods\cite{caselles1993geometric, li2010distance,vese2002multiphase}, proximal methods \cite{chambolle2011first}, etc have been developed over past few decades. The main technical difficulty herein is the non-differentiable nature of the original Mumford-Shah functional. To address this, the level-set approaches \cite{caselles1993geometric, li2010distance,vese2002multiphase} employ differentiable level function to obtain the segmentation mask using level-set evolutions. On the other hand, our method is a direct minimization of the Mumford-Shah functional thanks to the similarity between the softmax layer outputs and characteristic functions in Mumford-Shah functional. Moreover, this minimization can be done in a data-driven way using deep neural networks, which results in real-time computation.

Experimental results using various datasets verify that the Mumford-Shah functional can be synergistically combined to improve the segmentation performance.


\begin{figure*}[t!]
\centering
\includegraphics[width=18cm]{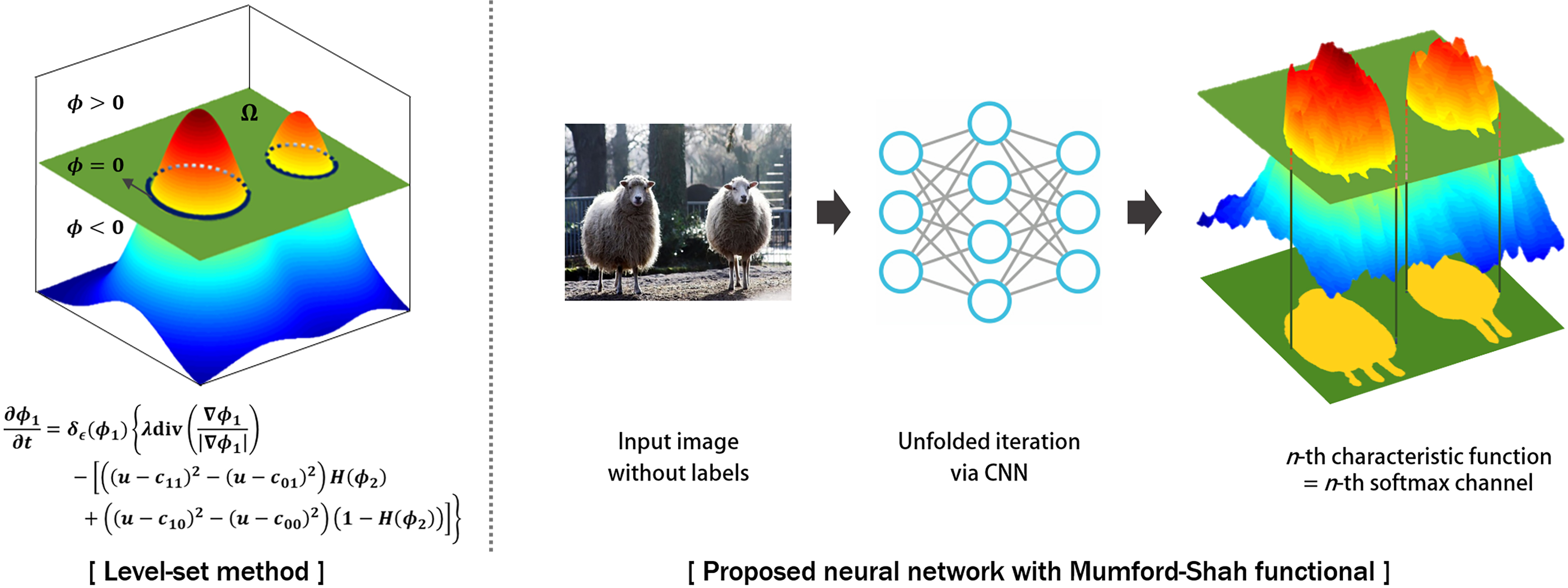}
\caption{High level comparison with the level-set method and the proposed method.}
\label{fig:concept}
\end{figure*}

\section{Related Works}
\subsection{Variational Image Segmentation}

Classical variational approaches consider the image segmentation problem as a clustering problem. Representative examples include k-mean \cite{dhanachandra2015image, chen1998image}, mean shift \cite{comaniciu2002mean, beck2016distributed}, normalized cuts \cite{cour2005spectral}, graph cuts \cite{boykov2001interactive}, and level-set \cite{caselles1993geometric, li2010distance,vese2002multiphase}. These approaches usually minimize an energy function. For example, one of the most well-known energy function is so-called Mumford-Shah functional \cite{mumford1989optimal}.

Specifically, for a given image measurement ${u}(\rb), \rb\in \Omega \subset \Rd^2$, {Mumford-Shah energy functional for image segmentation to $N$ class piecewise constant regions is defined as \cite{mumford1989optimal}}: 
\begin{align}
\mathcal{L}_{MS}(\chi;{u})
=\mathcal{E}(\chi,{u}) + \lambda \sum\limits_{n=1}^N \int_\Omega |\nabla \chi_n(\rb)|d\rb ,
\label{eqn:MS} 
\end{align}
where 
\begin{align}
\mathcal{E}(\chi,{u}) = \sum_{n=1}^N \int_{\Omega} |{u}(\rb)-c_n |^2 \chi_n(\rb) d\rb,
\label{eqn:levelset1term}
\end{align}
and $\chi_n(\rb)$ denotes the characteristic function for the $n$-th class such that
\begin{eqnarray}\label{eq:chi}
\sum_{n=1}^N \chi_n(\rb) = 1, \quad \forall \rb\in \Omega,
\end{eqnarray}
and $c_n$ denotes the average pixel value given by
\begin{align}
c_n = \frac{\int_\Omega {u}(\rb)\chi_n(\rb) d\rb}{\int_\Omega \chi_n(\rb)d\rb} .
\label{eqn:levelset_cent}
\end{align}
For multichannel images such as color images, this energy function \eqref{eqn:MS} can be easily extended by defining ${u}(\rb)$ and $c_n$ as vectors composed of each channel values.

For images with intensity inhomogeneities, the pixels within the same region may vary drastically
so that they cannot be assigned to a single constant value. To address this, Li et al. \cite{li2011level} proposed a local intensity clustering method by relaxing the piecewise constant assumption as:
 \begin{eqnarray}
 {u}(\rb) \simeq b(\rb)\sum_{n=1}^N c_n\chi_n(\rb),
  \end{eqnarray}
where $b(\rb)$ is a slowing varying bias field. Then, the authors in \cite{li2011level} modified the term \eqref{eqn:levelset1term} to
\begin{align}
& \mathcal{E}(\chi, {u}, b) = \nonumber \\
& \int \left( \sum\limits_{\omega \in I} \int_{\Omega} K(\rb-\rb')\|{u}(\rb')-b(\rb)c_\omega\|^2 \chi_\omega(\rb') d\rb' \right) d\rb ,
\label{eqn:biaslevelset} 
\end{align}
where $K(\rb-\rb')$ represents a kernel function. In \cite{li2011level}, a truncated Gaussian function is used for the kernel function $K$. Since the bias field $b(\rb)$ is simultaneously estimated with the class label $\{c_n\}$ when minimizing the energy function, this method enables image segmentation with intensity inhomogeneity correction.

Unfortunately, Mumford-Shah functional \eqref{eqn:MS} and its extension with the bias are not differentiable due to the characteristic function $\chi_n$. In order to make the cost function \eqref{eqn:MS} differentiable, Chan and Vese \cite{vese2002multiphase} proposed the multiphase level-set methods. Specifically, consider a set $\Phi=[\phi_1\cdots \phi_p]$, where $\phi_i:\Omega \to \mathbb{R}$ denotes a level function from an image domain $\Omega$ to the real number that represents the height of the level-set. They also define the vector Heaviside function \cite{vese2002multiphase}:
\begin{align}
H(\Phi)=\begin{bmatrix} H(\phi_1) & \cdots & H(\phi_p) \end{bmatrix},
\end{align}
where $H(\phi)=1$ when $\phi>0$ or $H(\phi)=0$, otherwise. Using this definition, one can generate $N:=2^p$ possibilities of the vector Heaviside function values to represent $N$ distinct classes within $\Omega$.

This leads to the Euler-Lagrangian equation for the level function. For example, for the classification of $N=4$ levels with two level functions (i.e. $p=2$ and $N=\{00,01,10,11\}$), the Euler-Lagrangian equations for the level functions are given by:
\begin{eqnarray}\label{eq:euler}
\frac{\partial \phi_1}{\partial t} &=& \delta_\epsilon(\phi_1) \left\{ \lambda \mathrm{div} \left(\frac{\nabla \phi_1}{|\nabla \phi_1|}\right) \right. \\
&&-\left[((u-c_{11})^2-(u-c_{01})^2)H(\phi_2) \right. 
\notag \\
&&\left. +\left((u-c_{10})^2-(u-c_{00})^2)(1-H(\phi_2))\right]\right\}, \notag
\end{eqnarray}
\begin{eqnarray}\label{eq:euler2}
\frac{\partial \phi_2}{\partial t} &=& \delta_\epsilon(\phi_2) \left\{ \lambda \mathrm{div} \left(\frac{\nabla \phi_2}{|\nabla \phi_2|}\right) \right. \\
&&-\left[((u-c_{11})^2-(u-c_{10})^2)H(\phi_1) \right. 
\notag \\
&&\left. +\left((u-c_{01})^2-(u-c_{00})^2)(1-H(\phi_1))\right]\right\}, \notag
\end{eqnarray}
where $\delta_\epsilon$ denotes an approximation to the one-dimensional Dirac delta function, {originated from the derivative of Heaviside function}. Thanks to the high-dimensional lifting nature of the level functions, the multiphase level-set approach can successfully segment spatially separated regions of the same class.

\subsection{CNN-Based Image Segmentation}
\subsubsection{Supervised {Semantic Segmentation}}
When {semantic} pixel-wise annotations are available, deep neural network approaches have become the main workhorses for modern segmentation techniques thanks to their high performance and fast runtime complexity \cite{chen2018deeplab, kampffmeyer2016semantic}. Since fully convolutional networks (FCNs) \cite{long2015fully} generate output map with same size of the input, various deep learning methods using FCNs are studied for semantic image segmentation \cite{badrinarayanan2015segnet, krahenbuhl2011efficient, christ2016automatic, tran2016fully, ronneberger2015u}.

\subsubsection{Weakly/Semi-supervised {Semantic Segmentation}}
Although the supervised learning methods {provide high performance semantic segmentation}, labeling images with pixel-level annotations is difficult to obtain for enormous amount of data. To address this issue, Qi et al. \cite{qi2016augmented} develop a unified method of semantic segmentation and object localization with only image-level signals. Papandreou et al. \cite{papandreou2015weakly} present expectation-maximization (EM) method to predict segmentation maps by adding image-level annotated images. Hong et al. \cite{hong2015decoupled} propose a decoupled architecture to learn classification and segmentation networks separately using the data with image-level supervision and pixel-level annotations. Souly et al. \cite{souly2017semi} and Hung et al. \cite{Hung_semiseg_2018} present the generative adversarial learning method for image segmentation.

\subsubsection{Unsupervised {Segmentation}}
Unsupervised learning methods have been also studied to segment image data without any ground-truth segmentation masks \cite{pham2018scenecut, wang2017unsupervised}. Several recent works present unsupervised approaches using classical optimization-based algorithms as a constraint functions \cite{xia2017w, moriya2018unsupervised, kanezaki2018unsupervised}.

\subsection{Our Contribution}

The main goal of this paper is to utilize classical Mumford-Shah functional with the modern neural network approaches. This results in the following contributions.

\begin{itemize}
\item In contrast to the classical multiphase level-set approach \cite{vese2002multiphase}, which relies on computational expensive level function evolution by the Euler-Lagrangian equation, in the proposed method, Mumford-Shah functional is directly minimized using a neural network by back-propagation.

\item Unlike the existing weakly/semi-supervised segmentation \cite{qi2016augmented, papandreou2015weakly, hong2015decoupled, Hung_semiseg_2018}, the proposed algorithm does not require the weak-labeled supervision for unlabeled data, but still uses these unlabeled images as elements of the training data, thanks to the Mumford-Shah functional that depends on pixel statistics. Thus, the proposed method greatly alleviates the task of manual annotation. 

\item While the existing CNN-based unsupervised learning methods \cite{pham2018scenecut, wang2017unsupervised,xia2017w, moriya2018unsupervised, kanezaki2018unsupervised} usually require complex pre- and /or post-processing, this post-processing step is not necessary and the algorithm only requires an addition of Mumford-Shah loss functional to the existing CNN approaches.

\item
While the combination of level-set method with CNN segmentation can be found in Kristiadi \cite{kristiadi2017deep} and Le \cite{le2018reformulating}, these approaches train the networks in supervised or weakly supervised manner, after which the level-set method is used as a refinement step of segmentation map. In contrast, our algorithm directly minimizes the Mumford-Shah functional so that it can be used for network training under semi-, unsupervised- and fully supervised setting.

\item
When the proposed loss can be used as a data-adaptive regularized function in a fully supervised segmentation algorithm, our loss allows the network to better adapt to the specific image statistics to further improve segmentation performance. 
\end{itemize}

\section{Theory}


\subsection{Key Observation}
One of the important observations in this paper is that the softmax layer in CNN can be used as a differentiable approximation of the characteristic function so that the direct minimization of Mumford-Shah functional is feasible.
This is different from classical multiphase level set methods, which obtain a differentiable energy function by approximating the characteristic function using a vector Heaviside function of multiphase level-sets.

Specifically, the $n$-th channel softmax output from a neural network is defined as follows:
\begin{eqnarray}\label{eq:softmax}
y_n(\rb) = \frac{e^{z_n(\rb)}}{\sum_{i=1}^N e^{z_i(\rb)}}, \quad n=1,2, \cdots, N ,
\end{eqnarray}
where $\rb\in \Omega$, and $z_i(\rb)$ denotes the network output at $\rb$ from the preceding layer before the softmax. The output value of \eqref{eq:softmax} is close to 1 when the pixel value at $\rb$ belongs to the class $n$. Furthermore, it is easy to show that 
\begin{eqnarray}\label{eq:constraint}
\sum_{n=1}^N y_n(\rb)=1,\quad \forall \rb \in \Omega ,
\end{eqnarray}
which is basically identical to the characteristic function property in \eqref{eq:chi}. This similarity clearly implies that the softmax function output can work as a differentiable version of the characteristic function for the class membership. 

Accordingly, we propose the following CNN-inspired {Mumford-Shah functional as a loss function}:
\begin{align}\label{eqn:multilevelset_loss}
\mathcal{L}_{MScnn}&(\Theta;x) \\
=& \sum\limits_{n=1}^{N} \int_{\Omega} |x(\rb)-c_n |^2 y_n(\rb) d\rb 
 + \lambda \sum\limits_{n=1}^{N} \int_\Omega |\nabla y_n(\rb)|d\rb, \nonumber
\end{align}
where $x(\rb)$ is the input, $y_n(\rb):=y_n(\rb;\Theta)$ is the output of softmax layer in \eqref{eq:softmax}, and
\begin{align}
c_{n} := c_n(\Theta)= \frac{\int_\Omega x(\rb)y_n(\rb;\Theta) d\rb}{\int_\Omega y_n(\rb;\Theta)d\rb} \ ,
\label{eqn:levelset_cent_2}
\end{align}
is the average pixel value of the $n$-th class, where $\Theta$ refers to the learnable network parameters.

Note that \eqref{eqn:multilevelset_loss} is differentiable with respect to $\Theta$. Therefore, the loss function \eqref{eqn:multilevelset_loss} can be minimized by backpropagation during the training. Another interesting property of the loss function \eqref{eqn:multilevelset_loss} is that it provides a self-supervised loss for segmentation. This property is useful for unsupervised segmentation setup where the label-based supervised loss is not available. Furthermore, the new loss function leads to image segmentation based on the pixel value distribution that can be augmented even when semantic labels are available.

\subsection{Learning Minimizers of Mumford-Shah Functional}
For a given cost function \eqref{eqn:multilevelset_loss}, a standard method for variational approach is to solve the corresponding Euler-Lagrangian equation \cite{osher2006level}:
\begin{eqnarray}
-\lambda \nabla \cdot \left( \frac{\nabla y_n}{|\nabla y_n|}\right) + ( x(\rb)-c_n)^2 - \sum_{i\neq n} (x(\rb)-c_i)^2  = 0,
\end{eqnarray}
for all $n$, where the last term comes from the constraint \eqref{eq:constraint}. A corresponding fixed point iteration to obtain the solution of the Euler-Lagrangian equation is given by
\begin{align}\label{eq:unroll}
y_n & ^{k+1} = \\
& y_n^{k} + \eta^k \left(\lambda \mathrm{div} \left( \frac{\nabla y_n^k}{|\nabla y_n^k|}\right) 
+\sum_{i=1}^N (-1)^{\delta(n,i)}( x(\rb)-c_i^k)^2 \right), \nonumber 
\end{align}
where the superscript $k$ denotes the $k$-th iteration and $\delta{(n,i)}$ represents a discrete Dirac delta function, and $\eta^k$ is a step size.

Inspired by the Learned iterative soft-thresholding algorithm (LISTA) \cite{gregor2010learning}, our neural network can be interpreted as an unrolled iteration of \eqref{eq:unroll}, truncated to a fixed number of iterations. This concept is illustrated in Fig.~\ref{fig:concept} for unsupervised segmentation problems by comparing it with the level set method.

\begin{figure}[!bt]
\centering
\includegraphics[width=8.8cm]{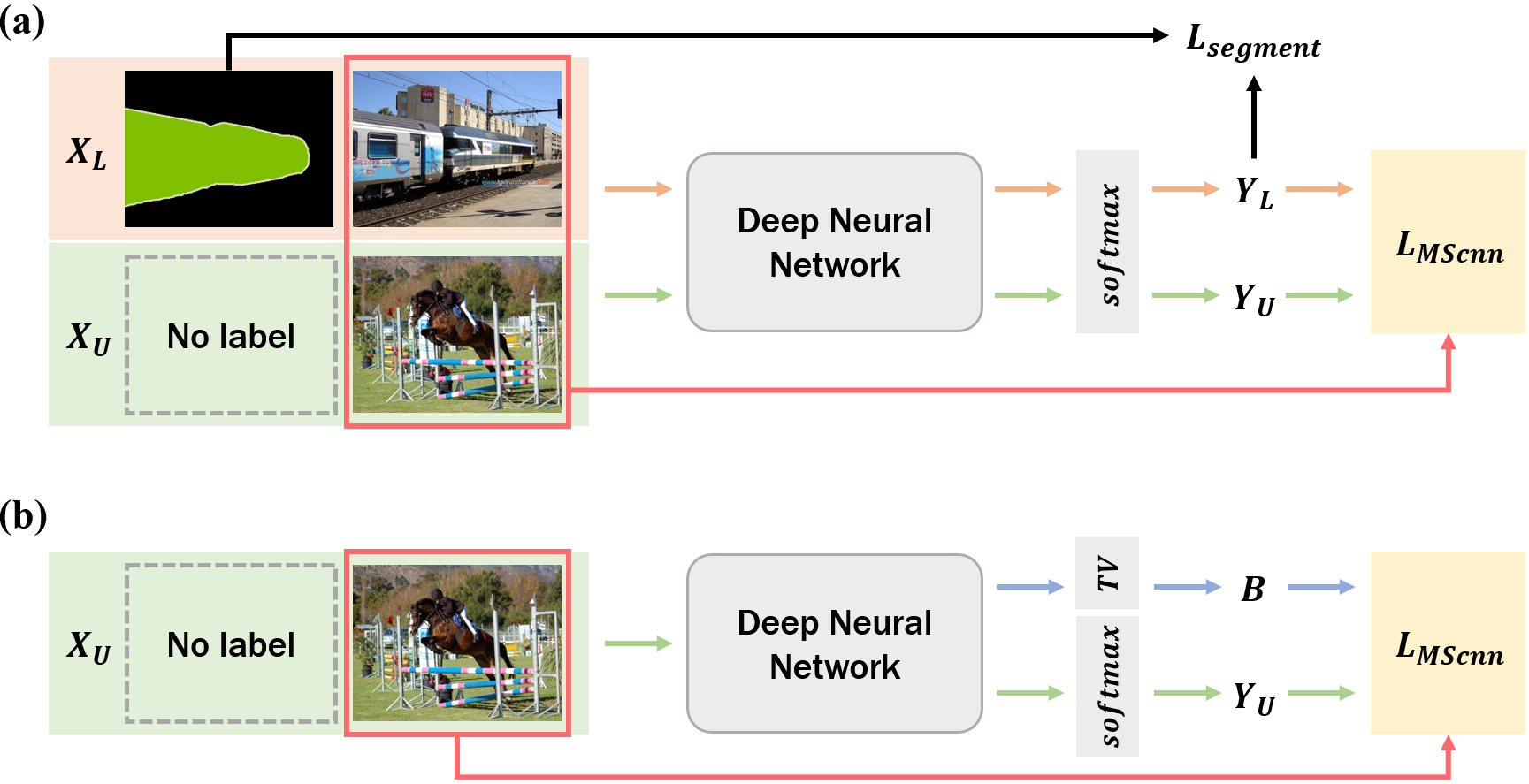}
\caption{The proposed image segmentation flow. 
(a) Semi-supervised segmentation: A deep neural network can take unlabeled images $X_U$ as well as the pixel-level annotated images $X_L$. For the input images $X_L$, the network can be trained using the segmentation loss $L_{segment}$ and Mumford-Shah loss functional $L_{MScnn}$. For the unlabeled data $X_U$, the network parameters are updated by minimizing $L_{MScnn}$. (b) Unsupervised segmentation: The neural network also estimates the bias field $B$ that is used for bias corrected Mumford-Shah loss functional.}
\label{fig:pipeline}
\end{figure}

\subsection{Application of Mumford-Shah Loss}

The new loss can be combined with any supervised, semi-supervised and unsupervised segmentation algorithm. Specifically, Fig. \ref{fig:pipeline}(a)(b) illustrates a typical use of the proposed loss function for the semi- and unsupervised segmentation tasks, respectively. Here, a deep neural network can be any existing segmentation network, which takes input images with or without pixel-level annotated data and generates segmentation maps. Only difference is the addition of the Mumford-Shah loss functional $L_{MScnn}$. Here, we describe the various application of the Mumford-Shah loss in more detail.

\subsubsection{{In the Presence of Semantic Labels}}
If the segmentation dataset has small amount of pixel-level semantic annotations, we apply the conventional segmentation loss function for pixel-level annotated data and our Mumford-Shah loss for unlabeled data so that the deep neural network can be trained without any estimated or weak supervisory labels (see Fig. \ref{fig:pipeline}(a)). 

Specifically, to use both labeled and unlabeled data, the loss function for network training is designed as following:
\begin{align} \label{eq:semi}
loss = \alpha \mathcal{L}_{segment} + \beta \mathcal{L}_{MScnn},
\end{align}
where 
\begin{align}
\alpha &=
 \left\{
  \begin{tabular}{ll}
    $1$, & if the input has labels \\ \nonumber
    $0$, & otherwise,
  \end{tabular}
 \right.
\end{align}
and $\beta$ is a hyper-parameter. $\mathcal{L}_{segment}$ is usually the cross-entropy function defined by:
\begin{align}
\mathcal{L}_{CE} = -\frac{1}{P}\sum\limits_{i=1}^{P} \sum\limits_{n=1}^{N} g_n(i)\, \mathrm{log}\, y_n(i),
\label{eq:crossentropy}
\end{align}
where $y_n(i)$ denotes the predicted probability of pixel $i$ belongs to class $n$ in the segmentation map, $g_n(i)$ denotes the semantic label, and $P$ is the number of pixels of input. Accordingly, when the network takes the input that does not have pixel-wise label, the network is trained only by the Mumford-Shah loss. Otherwise, the network is trained to segment specific regions and classify those categories using both the Mumford-Shah loss and the conventional segmentation loss. 

Meanwhile, in the presence of fully pixel-wise annotation labels, we set $\alpha=1$ during whole network training. However, in contrast to the existing supervised learning method that minimizes $\mathcal{L}_{segment}$, our neural network is trained using both $\mathcal{L}_{segment}$ and $\mathcal{L}_{MScnn}$ so that it can consider pixel similarity and semantic information to perform segmentation in more detail.

\subsubsection{{In the Absence of Semantic Labels}}
Since our loss {is a self-supervised loss}, the unsupervised image segmentation is also possible to estimate the cluster based on the pixel statistics by setting $\alpha=0$ in (\ref{eq:semi}). However, the Mumford-–Shah functional simply tries to force each segment to have similar pixel values with the contour length regularization, so that if the image pixel-values within each regions vary significantly due to the intensity inhomogeneities, it often produces separate segmented regions. Although the additional semantic loss can help to mitigate the problem, in unsupervised learning, such additional supervision is not possible. In this aspect, the bias field estimation \cite{li2011level} discussed above can provide additional supervision information.

Specifically, we impose the constraint that the total variation of the bias field is small so that we can choose the delta function kernel $K(\rb-\rb')=\delta(\rb-\rb')$ in \eqref{eqn:biaslevelset}. Then, \eqref{eqn:multilevelset_loss} can be modified as:
\begin{align}
\mathcal{L}_{MScnn}(\Theta;x) & = \sum\limits_{n=1}^{N} \int_{\Omega} |x(\rb)-b(\rb) c_n |^2 y_n(\rb) d\rb \nonumber \\
&+ \lambda \sum\limits_{n=1}^{N} \int_\Omega |\nabla y_n(\rb)|d\rb + \gamma \int_\Omega |\nabla b(\rb)|d\rb ,
\label{eqn:biaslevelset_loss}
\end{align}
where $\gamma>0$ is a fixed parameter and 
\begin{align}
c_{n} = \frac{\int_\Omega b(\rb) x(\rb)y_n(\rb) d\rb}{\int_\Omega b^2(\rb) y_n(\rb)d\rb} \ ,
\label{eqn:levelset_biascent}
\end{align}
which computes the centroid $c_n$ using the estimated bias in addition to the input image $x(\rb)$. Accordingly, the neural networks can simultaneously estimate the bias field and segmentation maps by minimizing \eqref{eqn:biaslevelset_loss}. 

The estimation of the bias field can be easily implemented in a neural network as shown in Fig.~\ref{fig:pipeline}(b). Specifically, in addition to $\{z_n(\rb)\}$ in \eqref{eq:softmax} that feeds into the softmax layers, the neural network is designed to output additional bias field $b_n(\rb)$. By a simple augmentation of bias field channel before the softmax layers and minimizing the modified loss function \eqref{eqn:biaslevelset_loss}, both bias and bias corrected segmentation mask can be obtained from the same neural network.

\section{Method}

\begin{figure*}[t!]
\centering
\includegraphics[width=18cm]{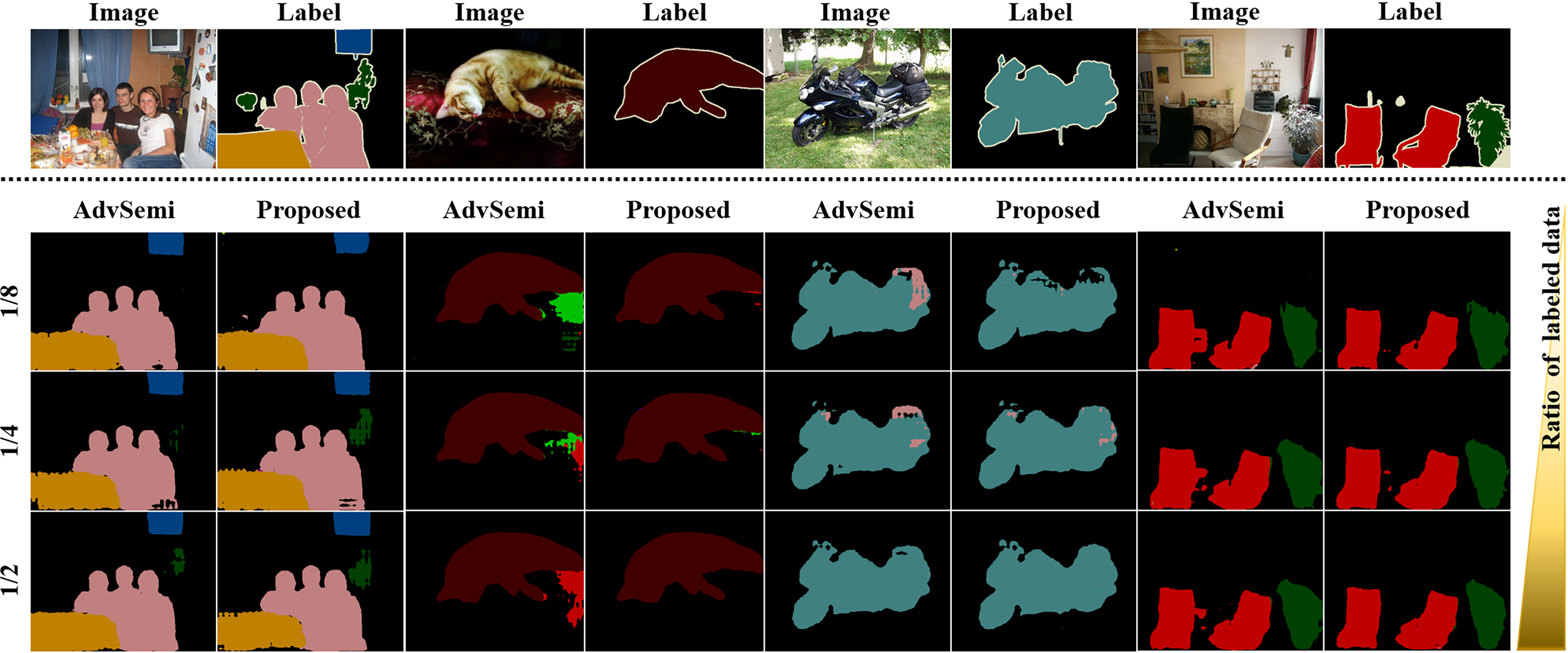}
\caption{Comparison with AdvSemi \cite{Hung_semiseg_2018} using the PASCAL VOC 2012 \textit{val} set. We randomly sampled 1/8, 1/4, 1/2 images as labeled data, and trained the models with both the labeled data and the rest of training images as unlabeled data. }
\label{fig:result_pascal248}
\end{figure*}

\subsection{Datasets}
\subsubsection{PASCAL VOC 2012}
To evaluate the semantic segmentation performance of our model on the daily captured natural images, we used PASCAL VOC 2012 segmentation benchmark \cite{everingham2010pascal}, consisting of 20 object classes. We also used the extra annotated images from Segmentation Boundaries Dataset (SBD) \cite{hariharan2011semantic} for training, so that we obtained an augmented training dataset of 10,582 images, and evaluated our semi-supervised algorithm on the validation set with 1449 images. 

\subsubsection{LiTS 2017}
To apply our method to medical image segmentation, we used Liver Tumor Segmentation Challenge (LiTS) dataset \cite{lits2017}. This provides 201 contrast-enhanced 3D abdominal CT scans and segmentation labels for liver and tumor regions with a resolution of $512 \times 512 $ pixels in each axial slices. However, among 201 data set, 70 scans do not have ground-truth labels. Accordingly, we only used 131 scans which were provided with pixel-wise semantic labels (i.e. liver and tumor). From the 131 scans provided with ground-truth labels, we used 118 scans for network training, and 13 scans for inference phase. We clipped the intensity values to the range [-124, 276] HU to ignore irrelevant details and normalized the images into $[0, 1]$ by dividing by the maximum value of data. Also, we downsampled the images into $256 \times 256$. 

\subsubsection{BRATS 2015}
The multimodal brain tumor image segmentation benchmark (BRATS) \cite{menze2015multimodal} contains 274 MR scans from different patients-histological diagnosis. Each scans has skull-stripped four MRI sequences: T1-weighted (T1), T1 with gadolinium enhancing contrast (T1c), T2-weighted (T2), and FLAIR. All training data with the size $240\times 240\times 155$ have the manual segmentation labels for several types of brain tumor. We trained the deep network for the complete tumor segmentation using 249 training data and evaluated our method on the 25 test set.

\subsubsection{BSDS500}
The Berkeley Segmentation Database (BSDS500) provides 200, 100, 200 color (RGB) images for the training, validation, and test images with human-annotated labels, respectively. We resized all training images to $512 \times 512 \times 3$.

\subsection{{Implementation Details}}
When pixel-level semantic labels exist in the datasets, we trained deep neural networks in a semi-supervised manner; otherwise, we trained the networks in an unsupervised setting. We implemented the all experiments using the pyTorch library in Python. 

\subsubsection{Semi-supervised Object Segmentation in Natural Images}
In order to evaluate the {proposed Mumford-Shah loss using} the PASCAL VOC 2012 dataset, we employed two different semi-supervised segmentation model: SSGAN \cite{souly2017semi} and AdvSemi \cite{Hung_semiseg_2018}. Then, we trained the networks with or without our loss in a semi-supervised setting. As a baseline network, we used a modified version of the DeepLab-v2 \cite{chen2018deeplab}.
This modified DeepLab-v2 does not use the multi-scale fusion, rather the Atrous Spatial Pyramid Pooling (ASPP) method is used in the last layer.

For fair comparison, we used same parameters between comparative methods and our method, except for the softmax-operated final output to compute our loss function. We randomly scaled and cropped the images into $321 \times 321$, and trained all models using two NVIDIA GeForce GTX 1080 Ti.

\subsubsection{Semi-supervised Tumor Segmentation in Medical Images}
For the tumor segmentation on LiTS 2017 and BRATS 2015, we used the modified U-Net \cite{kim2018cycle}. This modification comes from the pooling and unpooling layers of the U-Net \cite{ronneberger2015u} using the lossless decomposition: i.e. the four neighbor pixels of the input are decomposed into the four channels data with reduced size at the pooling layer, whereas the four reduced size channels are grouped together to an enlarged single channel at the pooling layer. The lossless modification improves the segmentation performance by retaining more details.

We stacked three adjacent slices in a volume as an input, and trained the network to generate segmentation masks corresponding to the center slice of the input. We used the Adam optimization method. For each LiTS and BRATS datasets, we set the initial learning rate as $10^{-5}$ and $10^{-4}$, and multiplied by 0.5 after every 20 epochs. Using a single GPU mentioned above, we trained the models with batch size 4 for 50 and 40 epochs on those datasets, respectively.

\subsubsection{Unsupervised Segmentation in Natural Images}
Recall that our method can convert the existing supervised segmentation network to unsupervised network by simply replacing the semantic loss with the proposed Mumford-Shah loss. To demonstrate this, we employed the U-Net \cite{ronneberger2015u} and trained the model without ground truth labels on the BSDS500 dataset. Additionally, in order to show that our loss function can improve the existing unsupervised segmentation network, we added the proposed loss function to the model of Backprop \cite{kanezaki2018unsupervised} and trained the network without labels. In the training of the U-Net, we used the Adam optimization algorithm with initial learning rate $10^{-4}$ and batch size 4. We stopped the training after 700 epochs. For the training of the modified Backprop \cite{kanezaki2018unsupervised}, we used SGD algorithm and initialized the parameters with Xavier initialization. We trained the network for 500 iterations and obtained the final segmentation maps.

Here, we also performed the comparative study using cnnLevelset \cite{kristiadi2017deep} which is one of the existing deep-learning-based level-set segmentation methods. Since this cnnLevelset uses images with a single object and weak bounding box labels for the prior step of segmentation, we trained our model on the PASCAL VOC 2012 dataset with the same condition for fair comparison. Specifically, we did not use the bounding-box labels in cost function for the segmentation steps, but used the labels only for pre-processing data and performed unsupervised learning for image segmentation. We implemented the U-Net architecture and used the Adam optimization algorithm with learning rate $10^{-4}$. We trained the network for 7200 iterations with batch size 16 on the PASCAL VOC 2012 training dataset using a single GPU.

\begin{table*}[t!]
  \centering
  \begin{tabular}{M{2.1cm}|M{0.2cm}M{0.25cm}M{0.25cm}M{0.25cm}M{0.25cm}M{0.2cm}M{0.2cm}M{0.2cm}M{0.2cm}M{0.3cm}M{0.2cm}M{0.2cm}M{0.2cm}M{0.25cm}M{0.25cm}M{0.25cm}M{0.25cm}M{0.2cm}M{0.2cm}M{0.3cm}M{0.4cm}|M{0.5cm}}
  \hline 
  \multicolumn{1}{c|}{\multirow{2}{*} {Methods}} & \multicolumn{21}{c|}{Class} & \multirow{2}{*} {mIoU} \\ 
   & bkg & aero & bike & bird & boat & bttle & bus & car & cat & chair & cow & tble & dog & hrse & mbk & prsn & plnt & shp & sofa & train & tv & \\ 
  \hline \hline
   \multicolumn{1}{l}{SSGAN \cite{souly2017semi}} \\ \hline
  \multicolumn{1}{l|}{\quad Baseline} & 0.82 & 0.28 & 0.09 & 0.27 & 0.03 & 0.17 & 0.12 & 0.30 & 0.35 & 0.05 & 0.04 & 0.05 & 0.29 & 0.11 & 0.23 & 0.42 & 0.10 & 0.26 & 0.04 & 0.09 & 0.18 & 0.205 \\   
  \multicolumn{1}{l|}{\quad SSGAN} & 0.88 & 0.66 & 0.23 & 0.66 & 0.42 & 0.51 & 0.79 & 0.70 & 0.74 & 0.23 & 0.53 & 0.33 & 0.67 & 0.53 & 0.57 & 0.69 & 0.34 & 0.58 & 0.32 & 0.68 & 0.48 & 0.549 \\
  \multicolumn{1}{l|}{\quad SSGAN + $\boldsymbol{\mathcal{L}_{MScnn}}$} & 0.90 & 0.72 & 0.29 & 0.60 & 0.42 & 0.58 & 0.82 & 0.66 & 0.76 & 0.18 & 0.58 & 0.40 & 0.59 & 0.56 & 0.45 & 0.71 & 0.35 & 0.61 & 0.29 & 0.64 & 0.59 & 0.558  \\   \hline 
     \multicolumn{1}{l}{AdvSemi \cite{Hung_semiseg_2018}} \\ \hline
  \multicolumn{1}{l|}{\quad Baseline} & 0.92 & 0.84 & 0.37 & 0.82 & 0.44 & 0.75 & 0.85 & 0.81 & 0.85 & 0.27 & 0.76 & 0.34 & 0.78 & 0.68 & 0.57 & 0.80 & 0.56 & 0.77 & 0.35 & 0.75 & 0.62 & 0.662 \\ 
  \multicolumn{1}{l|}{\quad AdvSemi} & 0.93 & 0.87 & 0.41 & 0.84 & 0.60 & 0.79 & 0.91 & 0.85 & 0.86 & 0.30 & 0.75 & 0.41 & 0.76 & 0.75 & 0.80 & 0.83 & 0.56 & 0.77 & 0.43 & 0.80 & 0.64 & 0.707 \\
  \multicolumn{1}{l|}{{\quad AdvSemi + $\boldsymbol{\mathcal{L}_{MScnn}}$}} & 0.93 & 0.85 & 0.41 & 0.85 & 0.60 & 0.78 & 0.91 & 0.83 & 0.86  & 0.28 & 0.76 & 0.43 & 0.80 & 0.75 & 0.79 & 0.83 & 0.52 & 0.81 & 0.42 & 0.78 & 0.74 & 0.711 \\
  \hline 
  \end{tabular} 
  \caption{Quantitative results with intersection-over-union (IoU) for all classes on the PASCAL VOC 2012 \textit{val} set using 1/4 labeled data. Baseline denotes the fully supervised methods with the same amount of labeled data.}
  \label{table:pascal_quan_class}
\end{table*}

\begin{table}[t!]
  \centering
  \begin{tabular}{M{2.7cm}|M{1.2cm}M{1.2cm}M{1.2cm}}
  \hline 
  \multicolumn{1}{c|}{\multirow{2}{*} {Methods}} & \multicolumn{3}{c}{Evaluation metrics} \\ 
   & IoU & Accuracy & Recall \\ 
  \hline  \hline
  \multicolumn{1}{l}{SSGAN \cite{souly2017semi}} \\ \hline
    \multicolumn{1}{l|}{\quad Baseline} & 0.205  & 0.259 & 0.719  \\
 \multicolumn{1}{l|}{\quad SSGAN} & 0.549  & 0.700 & 0.717  \\
  \multicolumn{1}{l|}{{\quad SSGAN + $\boldsymbol{\mathcal{L}_{MScnn}}$}} & 0.558  & 0.672 & 0.759  \\ \hline
  \multicolumn{1}{l}{AdvSemi \cite{Hung_semiseg_2018}} \\ \hline
  \multicolumn{1}{l|}{\quad Baseline} & 0.662 & 0.740 & 0.848  \\  
  \multicolumn{1}{l|}{\quad AdvSemi} & 0.707  & 0.794 & 0.858  \\
  \multicolumn{1}{l|}{{\quad AdvSemi + $\boldsymbol{\mathcal{L}_{MScnn}}$}} & 0.711 & 0.797 & 0.854  \\
  \hline 
  \end{tabular} 
  \caption{Quantitative comparison results with respect to three metrics (IoU, accuracy, and recall) on the PASCAL VOC 2012 \textit{val} set using 1/4 labeled data.}
  \label{table:pascal_quan_metric}
\end{table}

\section{Experimental Results}

\subsection{Semi-Supervised Object Segmentation in Natural Images} 

\subsubsection{Experimental Scenario}
Using randomly chosen 1/4 of whole images as labeled data, we trained all methods in a semi-supervised setting on PASCAL VOC 2012 segmentation benchmark. Also, to verify the performance of our method even with a small amount of labeled data, we evaluated our methods with AdvSemi \cite{Hung_semiseg_2018} using the 1/2, 1/4, and 1/8 labeled data. In this experiment, $\mathcal{L}_{segment}$ in (\ref{eq:semi}) was set as the original loss function in the comparative methods, and $\beta$ hyper-parameter for the proposed loss function was set to $10^{-7}$. For all semi-supervised learning, we did not estimate the bias field, since the partial semantic labels can help mitigate the problem from pixel intensity inhomogeneities. 

\subsubsection{Qualitative Evaluation}
Fig. \ref{fig:result_pascal248} illustrates the segmentation results from our proposed method over different amounts of training data. It shows that the objects segmentation performance using the proposed loss function becomes more accurate compared to AdvSemi \cite{Hung_semiseg_2018}, with increasing rate of labeled data for training. Also, even in the cases of the small number of labeled data, we can confirm that the objects are segmented in the direction of correct answer thanks to the pixel similarity information from our Mumford-Shah loss. 
 
\subsubsection{Quantitative Evaluation}
Table \ref{table:pascal_quan_class} and Table \ref{table:pascal_quan_metric} show the evaluation results on the PASCAL VOC 2012 \textit{val} dataset for the baseline comparative methods and our proposed methods. We used three evaluation metrics: mean of intersection-over-union (IoU), accuracy, and recall on the number of classes. Table \ref{table:pascal_quan_class} shows that the proposed method improved most of 20 classes in PASCAL VOC 2012. Specifically, by adding the proposed Mumford-Shah loss to SSGAN, our method improves the segmentation performance except for 6-7 classes. These tendency were also similarly appeared when combined with AdvSemi. In particular, we observed that certain three classes (car, chair, motorbike) did not improved, but in the classes of cow, table, sheep, and tv-monitor, the IoU scores were increased with the proposed method. Moreover, as shown in Table \ref{table:pascal_quan_ratio}, the proposed method using the Mumford-Shah loss improves performance with small labeled data compared to the existing network without the proposed loss.

\begin{table}[t!]
  \centering
  \begin{tabular}{M{2.7cm}|M{0.8cm}M{0.8cm}M{0.8cm}M{0.8cm}}
  \hline 
  \multirow{2}{*} {Methods} & \multicolumn{4}{c}{Ratio of labeled data amount} \\  
   & 1/8 & 1/4 & 1/2 & Full \\  
  \hline \hline
  \multicolumn{1}{l|}{DeepLab-v2 (baseline)} & 0.636 & 0.662 & 0.679 & 0.702 \\  
  \multicolumn{1}{l|}{AdvSemi \cite{Hung_semiseg_2018}} & 0.689 & 0.707 & 0.729 & N/A \\
  \multicolumn{1}{l|}{{AdvSemi + $\boldsymbol{\mathcal{L}_{MScnn}}$}} & 0.693 & 0.711 & 0.720 & 0.736 \\
  \hline 
  \end{tabular} 
  \caption{mIoU value comparison  for various semi-supervised learning method according to the ratio of labeled data amount on the PASCAL VOC 2012 \textit{val} set.}
  \label{table:pascal_quan_ratio}
\end{table}

\begin{figure*}[t!]
\centering
\includegraphics[width=18cm]{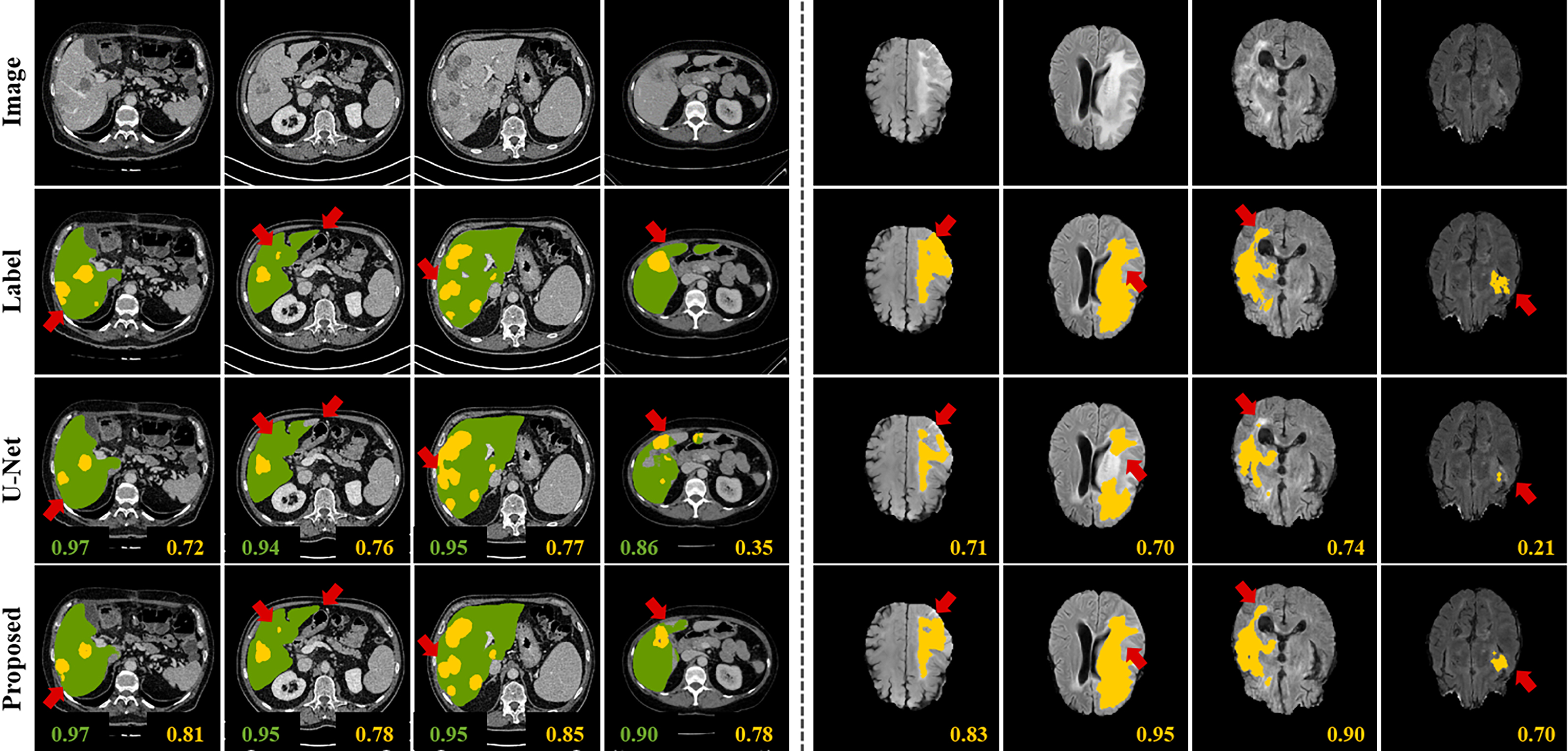}
\caption{Qualitative comparison results of the semi-supervised tumor segmentation. Left: results from our method using 1/10 labeled data on the LiTS dataset (green: liver, yellow: tumor). Right: results from our method using 1/4 labeled data on the BRATS dataset (yellow: tumor). First row: center slices of input. Second row: ground-truth labels. Third row: results from the supervised learning method \cite{ronneberger2015u} by $\mathcal{L}_{CE}$. Fourth row: results from our semi-supervised learning algorithm using $\mathcal{L}_{CE}$ and $\mathcal{L}_{MScnn}$. The scores on the bottom of each results denote Dice score.}
\label{fig:result_med}
\end{figure*}

\subsection{Semi-Supervised Tumor Segmentation in Medical Images} 

\subsubsection{Experimental Scenario}
In order to segment liver lesion in CT images effectively on the LiTS dataset, we trained the two modified U-Nets \cite{kim2018cycle}; the first is to segment a liver from CT abdominal scans, and the second is for lesion segmentation from the segmented liver. For the both models, we applied the semi-supervised learning scheme. Specifically, we randomly chose 1/23, 1/10, 1/3 scans of training data for the labeled data, and trained the two networks using both labeled and unlabeled data. At the inference phase, we evaluated the liver lesion segmentation performance on the validation set using these two segmentation networks trained in a semi-supervised manner.

For the tumor segmentation using BRATS dataset, we also used the modified U-Net using all three adjacent slices for each MR sequences (T1, T1c, T2, and FLAIR images) as an input. Among the training dataset, we randomly sampled 1/4 MR scans of the full data as labeled data for the semi-supervised learning scheme. Here, using the BRATS dataset, we conducted additional experiments on the hyper-parameter $\beta$ in (\ref{eq:semi}) by changing its value under the same training condition.

To verify the role of the proposed loss as a segmentation function, we compared our method to the supervised learning method by setting $\mathcal{L}_{segment} = \mathcal{L}_{CE}$, and we set the $\beta$ hyper-parameter as $10^{-6}$ and $10^{-7}$ for LiTS and BRATS dataset, respectively.

\subsubsection{Qualitative Evaluation}
Fig. \ref{fig:result_med} illustrates predicted liver lesion segmentation maps from CT slices and tumor segmentation maps from MR scans. These results verify that the Mumford-Shah loss enables the network to detect boundary of tumor region in more detail. {Also, there are several cases that the original deep network produces the segmentation map with the IoU score less than 0.5 but with the same network our proposed loss function improves the results with the IoU score of higher than 0.7.} Since the Mumford-Shah loss functional is computed with pixel-level information, we observed that tiny and thin tumors, which are hard to be distinguished with surrounding area, can be clearly segmented in our method.

\subsubsection{Quantitative Evaluation}
\begin{figure}[t!]
\centering
\includegraphics[width=8.7cm]{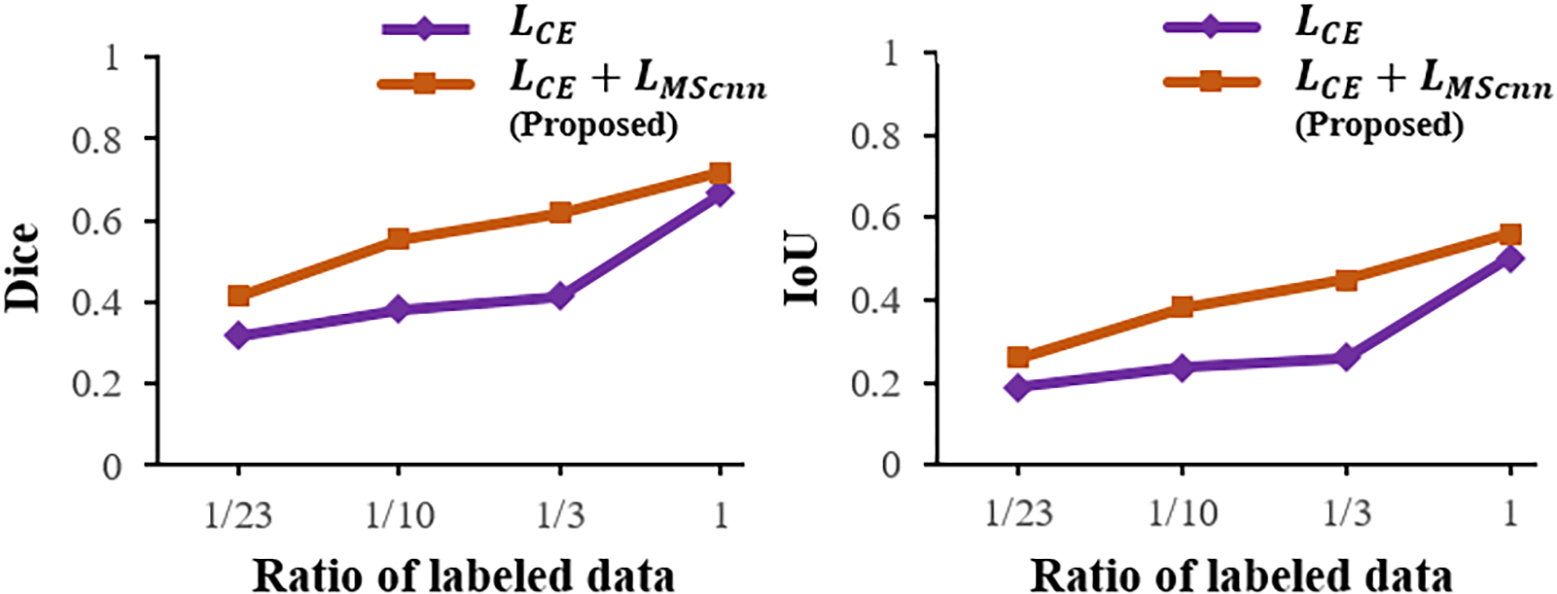}
\caption{Semi-supervised liver lesion segmentation performance by the proposed method using the validation set from LiTS 2017. The comparison was performed using various ratio of labeled data. Left: Dice score. Right: Intersection of Union (IoU) score.}
\label{fig:lits_quan}
\end{figure}

We evaluated the performance of tumor segmentation using the Dice coefficient, precision, recall, and intersection-over-union (IoU). Fig. \ref{fig:lits_quan} shows the scores on the 13 LiTS validation set according to the ratio of labeled and unlabeled data. The semi-supervised learning method using our proposed Mumford-Shah loss on the LiTS data brings significant performance improvement from $10\%$ to $20\%$, over different amounts of labeled data. 
Table \ref{table:brats_quan} also shows that the overall scores from our semi-supervised method on the brain tumor segmentation are higher than the general supervised learning method under the condition of 1/4 labeled data of training set. From these results, we confirmed that our loss function plays a role of regularizer and improves the segmentation performance, even when training the network using all training dataset with labels (i.e. ratio of labeled data = 1).

\begin{table}[t!]
  \centering
  \begin{tabular}{M{2.5cm}|M{0.7cm}M{1.0cm}M{0.7cm}M{0.7cm}}
  \hline 
  \multirow{2}{*} {Methods} & \multicolumn{4}{c}{Evaluation metrics} \\ 
  & Dice & Precision & Recall & IoU \\ 
  \hline \hline
  \multicolumn{1}{l|}{Modified U-Net} & 0.866 & 0.927 & 0.813 & 0.764 \\ 
   \multicolumn{1}{l|}{{Modified U-Net + $\boldsymbol{\mathcal{L}_{MScnn}}$}} & 0.880 &0.904 & 0.858 & 0.786 \\
  \hline 
  \end{tabular} 
  \caption{Quantitative comparison results of semi-supervised brain tumor segmentation on 25 validation scans of the BRATS dataset. We test our model with 1/4 training labeled data. }
  \label{table:brats_quan}
\end{table}

\begin{figure*}[t!]
\centering
\includegraphics[width=18cm]{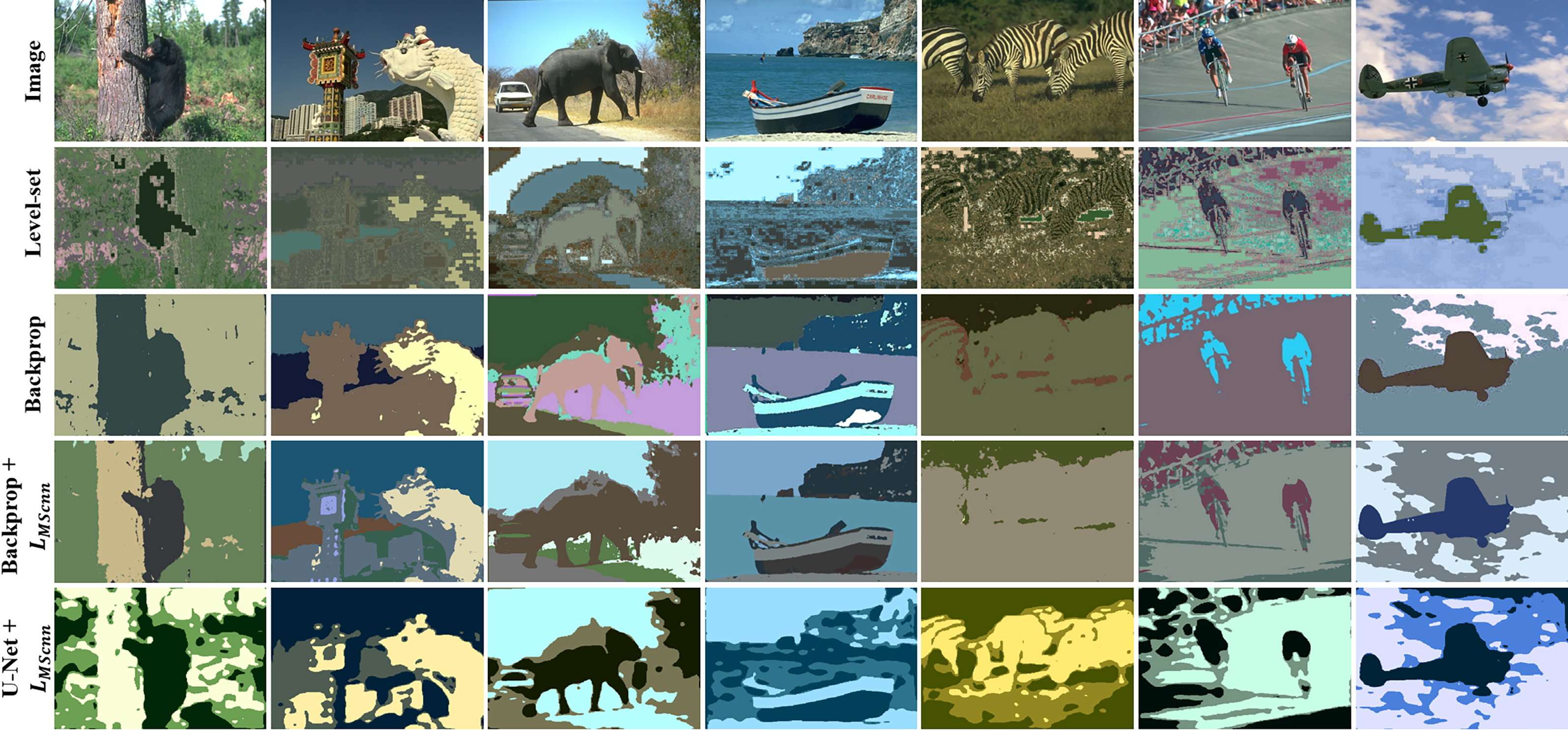}
\caption{Unsupervised image segmentation results using BSDS500 datasets. (First row) input images. (second row) level-set method, (third row) Backprop \cite{kanezaki2018unsupervised}, (fourth row) Improved Backprop using the proposed loss function, and (fifth row) U-Net with the proposed loss function.}
\label{fig:result_bsds}
\end{figure*}

\subsubsection{{Hyper-parameter Analysis}}
We also evaluated the sensitivity of the results on $\beta$ hyper-parameter using BRATS dataset as shown in Table \ref{table:brats_hyper}. In terms of Dice score, the performance improvement with the Mumford-Shah loss functional is about the same for various values of non-zero $\beta$. Among them, when $\beta$ is set to $10^{-7}$, the proposed method achieves the best Dice and IoU scores of $88\%$ and $79\%$, respectively.

\subsection{Unsupervised Segmentation in Natural Images}
\subsubsection{{Experimental Scenario}}
The goal of our unsupervised learning experiments is to directly verify that the combination of level-set framework and deep neural network improves the performance of semantic segmentation. Unlike the semi-supervised segmentation experiment, we incorporated the bias estimation, since it improves the segmentation performance for images with complicated foregrounds. We set the hyper-parameter $\beta$ as $10^{-7}$ for modified Backprop \cite{kanezaki2018unsupervised} with the proposed loss function.

\begin{table}[t!]
  \centering
  \begin{tabular}{M{1.5cm}M{0.8cm}|M{0.9cm}M{1cm}M{0.9cm}M{0.9cm}}
  \hline 
  \multirow{2}{*} {Data amount} & \multirow{2}{*} {$\beta$} & \multicolumn{4}{c}{Evaluation metrics} \\ 
  & & Dice & Precision & Recall & IoU \\ 
  \hline \hline
  1/4 & 0      & 0.867 &0.927 & 0.813 & 0.764 \\ 
  1/4 & $10^{-5}$  & 0.873 &0.901 & 0.847 & 0.775 \\
  1/4 & $10^{-6}$  & 0.878 &0.886 & 0.872 & 0.783 \\
  1/4 & $10^{-7}$  & 0.880 &0.904 & 0.858 & 0.786 \\
  1/4 & $10^{-8}$  & 0.877 &0.915 & 0.842 & 0.780 \\
  \hline 
  \end{tabular} 
  \caption{Sensitivity to $\beta$ for 25 validation scans of the BRATS dataset using 1/4 training labeled data.}
  \label{table:brats_hyper}
\end{table}

\subsubsection{Qualitative Evaluation on BSDS500}
Fig. \ref{fig:result_bsds} shows visual comparisons of the segmentation results. The proposed Mumford-Shah loss not only improves the performance of the existing methods (Backprop \cite{kanezaki2018unsupervised}) but also converts the supervised learning algorithms (U-Net) to an unsupervised learning method with the state-of-the art performance. We confirmed that our method enables segmentation of various foreground objects such as human and animals with higher performance than the comparative methods.

\subsubsection{{Quantitative Evaluation on BSDS500}}
Table \ref{table:bsds_quan} shows the results of comparisons and our method. We computed the Region Covering (RC), Probabilistic Rand Index (PRI), and Variation of Information (VI) as evaluation metrics. Our U-Net with the proposed loss function outperformed the conventional level-set \cite{vese2002multiphase} by $7 \%$ gain in RC, which verifies that the deep neural network enables improved object segmentation without ground-truth label. Also, from the comparison with the original Backprop, the modified Backprop with the Mumford-Shah loss achieved $6 \%$ gain in RC. These results confirmed that the proposed loss functional improves the unsupervised segmentation performance.

\begin{table}[t!]
  \centering
  \begin{tabular}{M{3.5cm}M{0.8cm}M{0.8cm}M{0.8cm}}
  \hline 
  \multirow{2}{*}{Method} & \multicolumn{3}{c}{Evaluation metrics} \\ 
   & {RC} & PRI & VI \\ 
  \hline\hline
  \multicolumn{1}{l}{Level-set \cite{vese2002multiphase} ($\#$regions=7)	}	    	& 0.403 & 0.703 & 2.722 \\ 
  \multicolumn{1}{l}{Backprop \cite{kanezaki2018unsupervised}} & 0.428 & 0.717 & 3.193 \\
    \hline
  \multicolumn{1}{l}{{Backprop + $\boldsymbol{\mathcal{L}_{MScnn}}$}} & 0.485 & 0.714 & 2.195 \\
 \multicolumn{1}{l} {{U-Net + $\boldsymbol{\mathcal{L}_{MScnn}}$}} 	& {0.470} & {0.727} & {2.322} \\
  \hline 
  \end{tabular} 
  \caption{Quantitative comparison results of unsupervised image segmentation on the BSDS500.}
  \label{table:bsds_quan}
\end{table}

\begin{figure*}[t!]
\centering
\includegraphics[width=16cm]{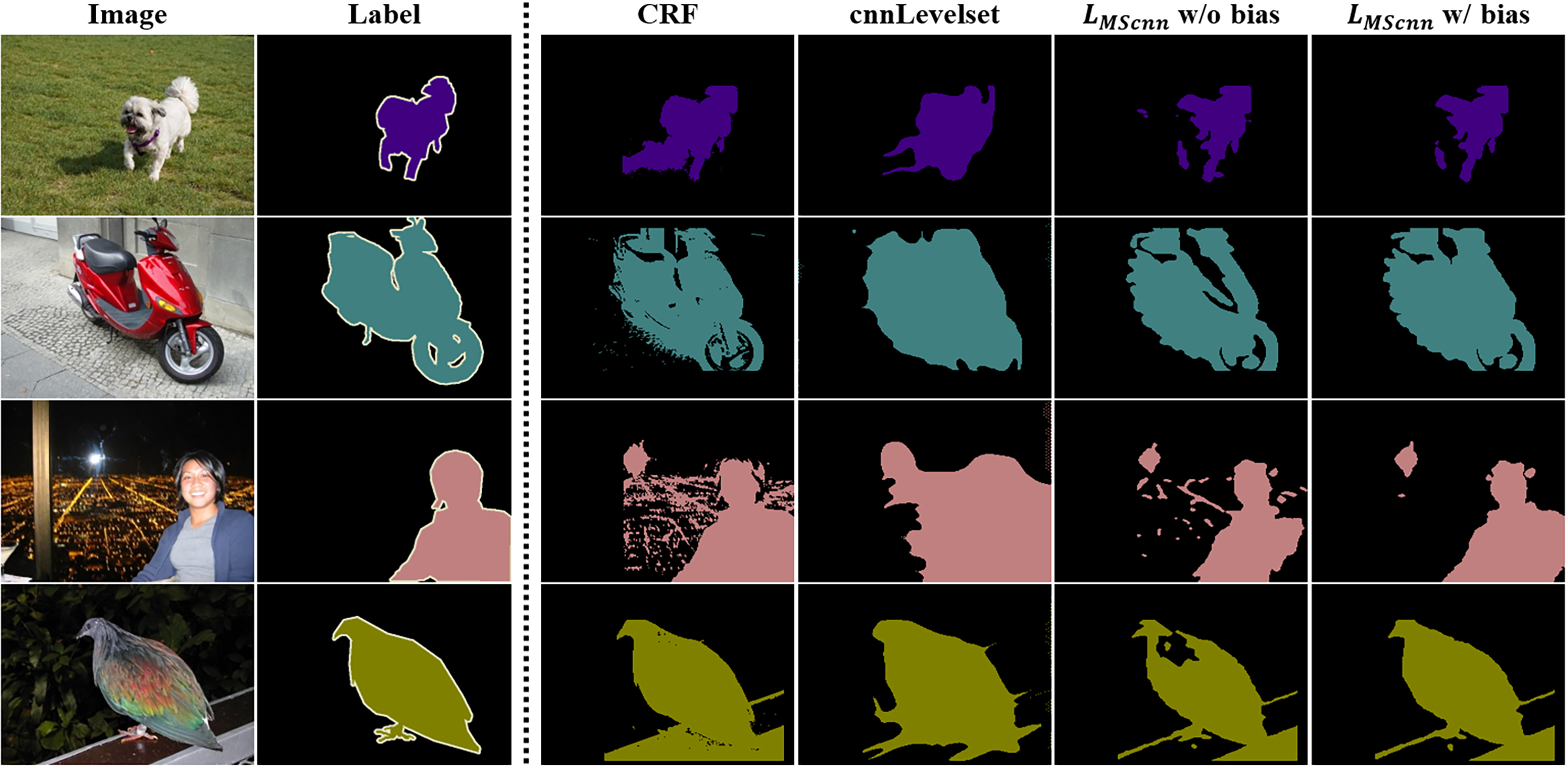}
\caption{Qualitative comparison results with cnnLevelset \cite{kristiadi2017deep} and CRF \cite{lafferty2001conditional}. The proposed method is implemented with and without bias field correction to investigate the effect of bias field estimation.}
\label{fig:result_pascal_levelcomp}
\end{figure*}

\subsubsection{{Comparison with CRF and CNN-based Level-set Segmentation Method}}
Table \ref{table:cnnlevelsetComp} and Fig. \ref{fig:result_pascal_levelcomp} show the comparison results with
the conventional CRF segmentation method \cite{lafferty2001conditional} and the existing deep-learning-based level-set method called cnnLevelset \cite{kristiadi2017deep}. As shown in Figure \ref{fig:result_pascal_levelcomp}, cnnLevelset generates very large segmentation with less detail, and CRF produces detailed edge information which is far from object semantics. On the other hand, the proposed method provides right balance between the
two approaches producing segmentation maps more related to the object semantics. We also found that our proposed algorithm produced segmentation maps with significantly smaller computational time. Moreover, we can observe that the bias field estimation helps in the segmentation of the complicated foreground object in images by modifying unnecessary details.

\begin{table}[t!]
  \centering
  \begin{tabular}{M{2.5cm}M{0.8cm}M{1.0cm}M{0.9cm}M{1.0cm}}
  \hline 
  \multirow{2}{*} {Methods} & \multicolumn{4}{c}{Evaluation metrics} \\ 
  & IoU & Accuracy & Recall & Time(sec) \\ 
  \hline \hline
   \multicolumn{1}{l}{{CRF \cite{lafferty2001conditional}}} & 0.346 &0.597 & 0.491 & 0.645 \\
   \multicolumn{1}{l}{cnnLevelset \cite{kristiadi2017deep}} & 0.349 & 0.532 & 0.552 & 0.215 \\  \hline
   \multicolumn{1}{l}{{$\boldsymbol{\mathcal{L}_{MScnn}}$ w/o bias}} & 0.351 & 0.542 & 0.539 & 0.014 \\ 
   \multicolumn{1}{l}{{$\boldsymbol{\mathcal{L}_{MScnn}}$} w/ bias} & 0.354 & 0.553 & 0.533 & 0.014 \\
  \hline 
  \end{tabular} 
  \caption{Quantitative comparison results of unsupervised object segmentation on PASCAL VOC 2012 \textit{val} set.}
  \label{table:cnnlevelsetComp}
\end{table}

\section{Conclusions}
In this paper, we proposed a novel Mumford-Shah loss functional and its variant for image segmentation using deep neural networks in semi-supervised and unsupervised manners. The main motivation for the new loss function was the novel observation that the softmax layer output has striking similarity to the characteristic function for Mumford-Shah functional for image segmentation so that the Mumford-Shah functional can be minimized using a neural network. Thanks to the self-supervised nature of the loss, a neural network could be trained to learn the segmentation of specific regions with or without small labeled data. Experiments on the various datasets demonstrated the efficacy of the proposed loss function for image segmentation.



\begin{IEEEbiography}[{\includegraphics[width=1in,height=1.25in,clip,keepaspectratio]
{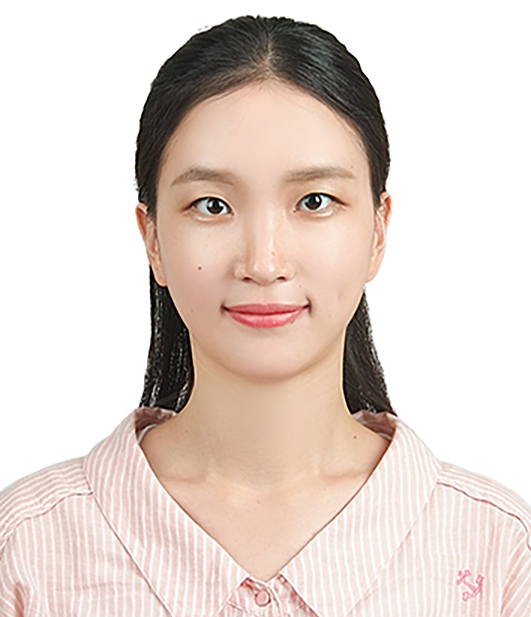}}]
{Boah Kim} received the B.S. and M.S. degrees from the Department of Bio and Brain Engineering, Korea Advanced Institute of Science and Technology (KAIST), Daejeon, Korea, in 2017 and 2019, respectively. She is currently a Doctoral program student at KAIST. Her research interests include machine learning for image processing such as image segmentation and registration in various applications.

\end{IEEEbiography}

\begin{IEEEbiography}[{\includegraphics[width=1in,height=1.25in,clip,keepaspectratio]
{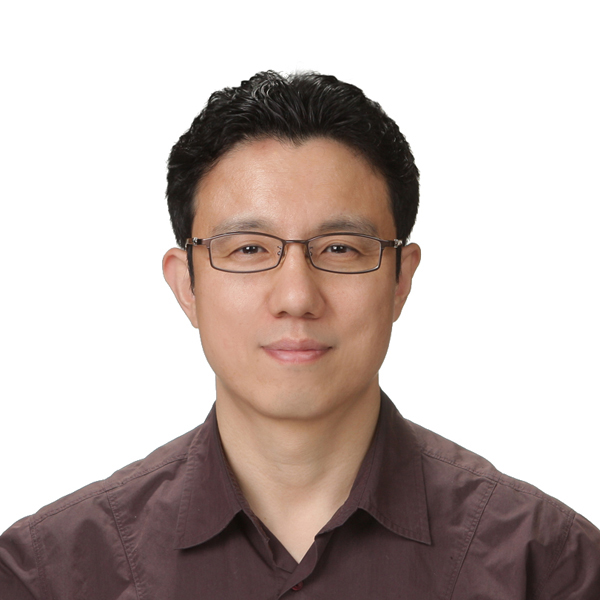}}]
{Jong Chul Ye} is currently a Professor of the Dept. of Bio/Brain Engineering and Adjunct Professor at Dept. of Mathematical Sciences of Korea Advanced Institute of Science and Technology (KAIST), Korea. He received the B.Sc. and M.Sc. degrees from Seoul National University, Korea, and the Ph.D. from Purdue University, West Lafayette.
From 1999 to 2001 he was a postdoc fellow at Coordindate Science Lab, University of Illinois at Urbrana-Champaign. From 2001 to 2004,  he worked at Philips Research and GE Global Research in New York. In 2004, he joined KAIST as an assistant professor, and he is now a Professor.  He has served as an associate editor of IEEE Trans. on Image Processing,  IEEE Trans. on Computational Imaging, Journal of Electronic Imaging,  an editorial board member for Magnetic Resonance in Medicine, and  an international advisory board member for Physics in Medicine and Biology. He is currently an associate editor for IEEE Trans. on Medical Imaging, a Senior Editor of IEEE Signal Processing Magazine, and a section editor of BMC Biomedical Engineering. He is  a vice-chair of IEEE SPS Technical Committee for Computational Imaging, and a general co-chair (with Mathews Jacob) for 2020 IEEE Symp. on Biomedical Imaging (ISBI), Iowa City. His group was the first place winner of the 2009 Recon Challenge at the ISMRM workshop, and the second place winner at 2016 Low Dose CT Grand Challenge organized by the American Association of Physicists in Medicine (AAPM). He was an advisor of best student paper awards (1st, and runner-up) at 2013 and 2016 IEEE Symp. on Biomedical Imaging (ISBI). His current research interests include machine learning and sparse recovery for various imaging reconstruction problems in x-ray CT, MRI, optics, ultrasound, etc.
\end{IEEEbiography}

\end{document}